\documentclass{article}
\usepackage{ijcai21}

\usepackage{times}
\usepackage{soul}
\usepackage{url}
\usepackage[hidelinks]{hyperref}
\usepackage[utf8]{inputenc}
\usepackage[small]{caption}
\usepackage{graphicx}
\usepackage{amsmath}
\usepackage{mathtools}
\usepackage{amsthm}
\usepackage{booktabs}
\usepackage{algorithm}
\usepackage{algorithmic}
\usepackage{amsmath,amssymb}
\usepackage{nicefrac}       % compact symbols for 1/2, etc.
\usepackage{multirow}
\usepackage{subfigure}
\usepackage{xspace}
\usepackage{comment}
\usepackage{bibentry} % for nobibliography (appendix only)
\usepackage{color}

\newtheorem{theorem}{Theorem}

\newtheorem{corollary}{Corollary}

\newcommand{\alname}{NKCVS\xspace}
\newcommand{\ealname}{E-NKCVS\xspace}
\DeclareMathOperator{\argmax}{argmax}
\DeclareMathAlphabet\mathbfcal{OMS}{cmsy}{b}{n}

\title{An Ensemble Noise-Robust K-fold Cross-Validation Selection Method \\ for Noisy Labels}

\author{
Yong Wen\thanks{Equal contribution.}\and
Marcus Kalander\footnotemark[1]\and
Chanfei Su\and
Lujia Pan
\affiliations
Noah's Ark Lab, Huawei Technologies
\emails
\{wenyong4,marcus.kalander,suchanfei1,panlujia\}@huawei.com
}

\begin{document}
\maketitle

\begin{abstract}
We consider the problem of training robust and accurate deep neural networks (DNNs) when subject to various proportions of noisy labels. Large-scale datasets tend to contain mislabeled samples that can be memorized by DNNs, impeding the performance. With appropriate handling, this degradation can be alleviated.
There are two problems to consider: how to distinguish clean samples and how to deal with noisy samples. 
In this paper, we present Ensemble Noise-robust $K$-fold Cross-Validation Selection (\ealname) to effectively select clean samples from noisy data, solving the first problem.
For the second problem, we create a new pseudo label for any sample determined to have an uncertain or likely corrupt label. \ealname obtains multiple predicted labels for each sample and the entropy of these labels is used to tune the weight given to the pseudo label and the given label.
Theoretical analysis and extensive verification of the algorithms in the noisy label setting are provided.
We evaluate our approach on various image and text classification tasks where the labels have been manually corrupted with different noise ratios. Additionally, two large real-world noisy datasets are also used, Clothing-1M and WebVision.
\ealname is empirically shown to be highly tolerant to considerable proportions of label noise and has a consistent improvement over state-of-the-art methods. Especially on more difficult datasets with higher noise ratios, we can achieve a significant improvement over the second-best model.
Moreover, our proposed approach can easily be integrated into existing DNN methods to improve their robustness against label noise.

\end{abstract}
\section{Introduction}

% Overall introduction to the problem of noisy label and what the challenges are
Together with the resurgence and remarkable success of DNNs, large-scale datasets have become increasingly common. 
For supervised learning tasks, modern DNNs generally require the datasets to be annotated with accurate labels to achieve high performance. However, to correctly label large amounts of data is very costly and error-prone, even high-quality hand-labeled benchmark dataset such as ImageNet~\cite{deng2009imagenet} contains mislabeled samples~\cite{northcutt2019confident}. There exist alternative, low-cost methods, including large-scale annotation through crowd-sourcing~\cite{sheng2008get} and online web queries~\cite{divvala2014learning}, but these inevitably yield a higher proportion of incorrect class labels. 

% Shortly about current approaches and some relevant papers
DNNs are prone to overfitting to corrupted data samples, which increases the generalization error of the network~\cite{zhang2017understanding}. To address this issue, numerous algorithms have been proposed to train DNNs in a way robust to label noise~\cite{wang2019sl,Xu2019L_DMIAI}. 
The capability of DNNs to fit noisy data has been further studied by Chen et al.~\shortcite{chen2019understanding}. They showed that, for symmetric noise, the test accuracy is a quadratic function of the noise ratio, and claim that generalization occurs in the sense of distribution. In this paper, we relax their assumptions and give a theoretical analysis of the impact that an imperfect classifier has. 
Our findings demonstrate that, while the noise level has a significant impact, the performance of the classifier is key. % add some other findings?

% Build upon k-fold cross-validation
Based on our analysis, we propose \ealname, a novel ensemble method based on $K$-fold cross-validation
%~\cite{stone1974cross}
to increase the generalization performance. We empirically evaluate our solution and demonstrate that it outperforms the state-of-the-art, proving the effectiveness of our method.
In summary, our contributions are as follows.
\begin{itemize}
    \item We propose a novel method (\ealname) based on a combination of $K$-fold cross-validation and ensemble learning. Samples are selected from the noisy data by keeping those where the predicted label matches the given (noisy) label. Any non-selected samples can then either be discarded or re-weighted to have a lower impact. Mixup~\cite{zhang2017mixup} is applied during training to augment the data.
    %Furthermore, we provide insights into how the algorithm works and give a theoretical analysis on how to parameterize the model. % optimize the results
    \item We further propose a label re-weighting scheme for samples that are likely erroneous. For these uncertain samples, we consider both the given label and a generated pseudo label with the weight set using the entropy of the predicted labels given by \ealname.
    \item We empirically show that the proposed solution outperforms state-of-the-art noise-robust methods on image recognition and text classification tasks on multiple datasets. % without the need for auxiliary clean labels during training.
    Moreover, our solution can easily be incorporated into existing network architectures to enhance their robustness to noisy labels.
\end{itemize}
\section{Related Work}
%Label noise is ubiquitous and inevitable in the real world. Training DNNs with noisy labels can significantly deteriorate their performance.
There have been numerous approaches proposed to deal with noisy labels. These can generally be categorized into three types.
% correct the corrupted labels
The most straightforward way is to improve the quality of a dataset by removing or correcting corrupted samples.
There have been multiple strategies proposed to identify the most likely corrupted samples, including using conditional random fields~\cite{vahdat2017cdf},
knowledge graphs distilling knowledge from noisy data~\cite{li2017distill}, and a label cleaning network to achieve noise-robust classifications~\cite{veit2017learning}. 
However, auxiliary clean data are often required and not always obtainable.

% Modify the loss function
Another approach is to reformulate the loss function.
Theoretical studies by Ghosh et al~\shortcite{ghosh2017robust} prove that the mean absolute error (MAE) is robust to label noise under certain assumptions. Inspired by Ghosh's work, other robust losses have been proposed. Ma et al.~\shortcite{ma2018d2l} correct the loss to avoid overfitting to noisy labels. Wang et al.~\shortcite{wang2019sl} propose symmetric cross-entropy learning by balancing cross-entropy and a noise-tolerant reverse cross-entropy while Zhang et al.~\shortcite{zhang2018gce} propose a set of noise-tolerant loss functions that generalize both the categorical cross-entropy and MAE.
Xu et al.~\shortcite{Xu2019L_DMIAI} introduce Determinant-based Mutual Information (DMI) loss which is a generalized version of mutual information and provably insensitive to instance-independent label noise.

Refinement of the training process has also been explored to deal with noisy labels.
MentorNet~\cite{jiang2017mentornet} is proposed to supervise the training of a student network and make it focus on samples with a higher probability of being labeled correctly.
Following the same idea, Co-teaching~\cite{han2018coteach} trains a network with the most confident samples as output by a second network. 
Meanwhile, DivideMix~\cite{li2020dividemix} fit a two-component mixture model to obtain the per-sample label confidence, then use this information to divide the training data into a labeled set and an unlabeled set. The semi-supervised technique MixMatch~\cite{berthelot2019mixmatch} is then applied for training. 
In a similar fashion, MentorMix~\cite{jiang2020beyond} also takes advantage of Mixup~\cite{zhang2017mixup} and merges it with MentorNet to minimize the empirical vicinal risk using curriculum learning.

%Apart from these, different augmentations to the network structure to model the label noise statistics have been proposed~\cite{goldberger2016layer,jindal2016dropout}.

Our proposed method refines the training process by adding $K$-fold cross-validation and an ensemble to deal with the noisy labels and adjust per-sample label weights during training. 
\section{Preliminaries}
\label{sec:preliminaries}
We consider the $Q$-class classification problem. Given a dataset $\mathcal{D} = \{\mathbf{x}_i, y_i\}_{i=1}^{n}$, where $\mathbf{x}_i \in \mathcal{X} \subset \mathbf{R}^d$ denotes the $i$-th sample in the $d$-dimensional space with its observed label as $y_i \in [Q] = \{1, 2, ...,Q\}$. The given label $y_i$ may be corrupt and we thus denote $y_i^{*}$ as the ground-truth label of sample $i$. A sample $(\mathbf{x}_i, y_i)$ is referred to as clean when it is labeled correctly, i.e., $y_i=y_i^{*}$. In this work, we examine two types of artificial noise, symmetric noise and asymmetric noise. We introduce a noise transition matrix $T \in \mathbf{R}^{Q\times Q}$, where $T_{jk} = P(y=k|y^{*}=j)$, to characterize the probability of samples in the $j$-th class being flipped to the $k$-th class label.
\newtheorem{myDef}{Definition}
\begin{myDef}
\label{def:sym-noise}
    (\textbf{symmetric noise}) Given noise ratio $\epsilon$, we define the noise transition matrix as $T_{jj} = 1 - \epsilon, j\in [Q]$, and $T_{jk} = \frac{\epsilon}{Q - 1}, k \neq j, k \in [Q]$.
\end{myDef}

\begin{myDef}
\label{def:asym-noise}
    (\textbf{asymmetric noise}) Given noise ratio $\epsilon$, $T_{jj} = 1 - \epsilon, j\in [Q]$, and $T_{jk} = \epsilon$, for some $k \neq j, k \in [Q]$,  $T_{jk} = 0$ otherwise.
\end{myDef}

\noindent
Both noise types make the common assumption that the noise is data-independent given the true class label, i.e., $P(y_i|y_i^*;\mathbf{x}_i) = P(y_i|y_i^*)$. Asymmetric (class-dependent) noise is designed to imitate real-world label noise which often arises due to annotators mistaking similar classes. This is simulated by flipping a fraction of a class's labels to a similar class (e.g., truck $\rightarrow$ automobile, cat $\rightarrow$ dog).

For simplicity and consistency, we denote the neural network classifier parameterized by $\theta$ as $f(\mathbf{x}, \theta)$, where $f$ is an element of a  functional space $\mathcal{F}$ which maps the feature space to the label space $f: \mathcal{X} \to \mathbf{R}^Q$. 
We further denote $P(y^f|\mathbf{x}, \theta)$ and $\hat{y} = \argmax_{i} P(y^f = i|\mathbf{x}, \theta)$ as the class probability distribution and the predicted label, respectively. The loss is denoted as $\mathcal{L}(f(\mathbf{x}, \theta), y)$, or $\mathcal{L}(\mathbf{x}, y)$ for short. Finally, the confusion matrix of classifier $f$ is denoted as $C \in \mathbf{R}^{Q\times Q}$, where $C_{jk} = P(\hat{y}=k|y^{*}=j)$.
% $C_{jk} = P(\hat{y}=k|y^{*}=j)$, $C \in \mathbf{R}^{Q\times Q}$.
%Denote the train dataset as $\mathcal{D}_{train}$, the validation dataset as $\mathcal{D}_{val}$ and the test dataset as $\mathcal{D}_{test}$.
% Already in definition 1
%, noise ratio as $\epsilon$.

\section{Ensemble Noise-Robust K-fold Cross-Validation Selection}
In this section, we present the details of our sample selection strategy to obtain clean samples and our re-weighting scheme for samples with uncertain or likely corrupted labels. Our goal is to select clean samples from the noisy dataset $\mathcal{D}$, and consecutively train a deep learning model with the selected samples and the re-weighted non-selected samples.
The optimal scenario would be to select all samples $(\mathbf{x}_i, y_i)$ where $y_i=y^*_i$ from $\mathcal{D}$ and re-weight all non-selected samples to use their correct label $y^*_i$ as the training label.

%Here we present the main idea and the proposed algorithms for selecting clean samples from noisy label data. 
To effectively filter out noisy samples, we present a Noise-robust $K$-fold Cross-Validation Selection (\alname) method in Algorithm~\ref{alg:nkcvs}. Following the standard $K$-fold cross-validation scheme, the dataset is randomly partitioned into $K$ equal-sized subsets ($D_1,D_2,...,D_K$) (line~2). 
The data is split into training data $D_*$ consisting of $K-1$ subsets and a single subset $D_j = D \setminus D_*$ (line~4). We augment the training data $D_*$ and train a DNN model with the standard cross-entropy loss (lines~5-6). The model is then used to predict the labels of all samples in $D_j$ (lines~7-9), and we select any samples where the predicted label $\hat{y}$ matches the given label $y$ (lines~10-11). 
This process is repeated $K$ times until all samples have been tested once.

The training data augmentation is done following Mixup~\cite{zhang2017mixup}. Each sample $(\mathbf{x}_1, y_1)$ is interpolated with another randomly chosen sample $(\mathbf{x}_2, y_2)$ from the same mini-batch. For each such pair of samples, a mixed sample $(\mathbf{x}', y')$ is computed by:

\begin{align}
\label{eq:mixup}
    \lambda &\sim Beta(\alpha, \alpha),\\
    \lambda' &= \max(\lambda, 1 - \lambda),\\
    x' &= \lambda'x_1 + (1-\lambda')x_2,\\
    y' &= \lambda'y_1 + (1-\lambda')y_2.
\end{align}

We further propose an extended ensemble version in Algorithm~\ref{alg:enkcvs} (\ealname). 
%In the spirit of ensembles, the original \alname is iterated $M$ times. Each iteration $i$ yields a separate set of samples $\mathcal{T}_i$. 
In ensemble learning, multiple predictions are combined to obtain better predictive performance. Following this idea, we iterate NKCVS $M$ times with each iteration $i$ yielding a separate set of samples $\mathcal{T}_i$. 
We finally select all samples $e \in \mathcal{D}$ that fulfill the condition,
\begin{equation} 
\label{eq:contains}
\sum\limits_{i=1}^{M} I(e \in \mathcal{T}_i) \geq t,
\end{equation}
where $0 < t \leq M$ is the threshold to retain $e$ and $I$ is an indicator function returning 1 if $e \in \mathcal{T}_i$, otherwise 0.

For each sample, we save all $M$ predicted labels in $\hat{\mathcal{Y}}$. These are used to create pseudo labels for all non-selected samples and adjust the weight between the pseudo and given labels as described in Section~\ref{subsec:re-weight}.

\subsection{Evaluation strategy and theoretical analysis}
\label{subsec:theoretical-analysis}
\renewcommand{\algorithmicrequire}{ \textbf{Input:}} 
\renewcommand{\algorithmicensure}{ \textbf{Output:}}
\begin{algorithm}[t!]
	\caption{Noise-Robust K-fold Cross-Validation Selection (\alname)}
	\label{alg:nkcvs}
	\begin{algorithmic}[1]
		\REQUIRE noisy dataset $\mathcal{D}$, number of splits $K$.
	    \STATE $\mathcal{SS} = \{\}$, $\hat{Y} = [\ ]$
		\STATE Split $\mathcal{D}$ into $\{\mathcal{D}_1, \mathcal{D}_2, ..., \mathcal{D}_K\}$
    	\FOR{$j=1$ {\bfseries to} $K$}
    		\STATE $\mathcal{D}_{*} = \mathcal{D} \setminus \mathcal{D}_j$
    		\STATE $\Tilde{\mathcal{D}}_* = \textit{mixup}(\mathcal{D}_*)$
    		\STATE Train $f(\cdot, \theta)$ with $\Tilde{\mathcal{D}}_{*}$ to obtain $\theta^{*}$
    		\FOR{$(\mathbf{x}, y) \in \mathcal{D}_j$}
    		    
    		    \STATE Predict labels $\hat{y}$ with $f(\mathbf{x}, \theta^*)$
    			\STATE Append $(\mathbf{x}, \hat{y})$  to $\hat{Y}$
    			\IF{ $y == \hat{y}$ }
    				\STATE $\mathcal{SS} = \mathcal{SS} \cup \{(\mathbf{x}, y)\}$
    			\ENDIF
    		\ENDFOR
    	\ENDFOR
		\ENSURE the selected sample set $\mathcal{SS}$, predicted label set $\hat{Y}$.
	\end{algorithmic}
\end{algorithm}
\begin{algorithm}[t!]
	\caption{Ensemble NKCVS (\ealname)}
	\label{alg:enkcvs}
	\begin{algorithmic}[1]
		\REQUIRE  noisy dataset $\mathcal{D}$, number of splits $K$, number of iterations $M$, threshold $t$.
		\STATE $\mathcal{SS} = \{\}$, $\hat{\mathcal{Y}} = \{\}$.
		\FOR{$i=1$ {\bfseries to} $M$}
		    \STATE Set $\mathcal{T}_i, \hat{\mathcal{Y}}_i \gets \textit{NKCVS}(D, K)$
		    %\STATE $\mathcal{T} = \mathcal{T} \cup {\mathcal{T}_i}$
    	\ENDFOR
    	\FOR{$e \in D$}
    	   %\STATE $c$ = occurrences of $e$ in $\mathbfcal{T}$
    	   %\STATE $c$ = occurrences of $e$ in $\{\mathcal{T}_1,...,\mathcal{T}_M\}$
    	   \STATE $c$ = $\sum_{i=1}^M I(e \in \mathcal{T}_i)$
    	   \IF{$c \geq t$}
    	       \STATE $\mathcal{SS} = \mathcal{SS} \cup \{e\}$
    	   \ENDIF
        \ENDFOR
		\ENSURE the selected sample set $\mathcal{SS}$, predicted label set $\hat{\mathcal{Y}}$.
	\end{algorithmic}
\end{algorithm}

\noindent
To evaluate the algorithms' ability to select clean samples, we adapt the standard definitions of precision and recall to our scenario while keeping the original intent behind the metrics intact. We denote the selected sample set as $\mathcal{SS}$, and define the clean samples $\mathcal{CS}$ and clean selected samples $\mathcal{CSS}$ as follows:
\begin{equation} \label{tp}
    \begin{aligned}
        \mathcal{CS}  &:= \{(\mathbf{x}, y) \in \mathcal{D}, y = y^{*}\},  \\
        \mathcal{CSS} &:= \{(\mathbf{x}, y) \in \mathcal{SS}, y = y^{*}\}.  
    \end{aligned}
\end{equation}
We measure the ability of identifying the clean samples using \textit{precision} and \textit{recall}, defined as:
\begin{equation} \label{eq:prec-rec}
    Precision     := \frac{|\mathcal{CSS}|}{|\mathcal{SS}|}, \ \ \  
    Recall        := \frac{|\mathcal{CSS}|}{|\mathcal{CS}|},
\end{equation}
where $|\cdot|$ denotes the number of samples in a set. Thus, \textit{precision} expresses the fraction of clean samples in $\mathcal{SS}$, 
while \textit{recall} represents the fraction of clean samples in $\mathcal{SS}$ over all clean samples in $\mathcal{D}$.
%while \textit{recall} represents the fraction of the total amount of  clean samples that were actually retrieved in $\mathcal{SS}$.
The performance of Algorithm~\ref{alg:enkcvs} is theoretically quantified in Theorem~\ref{the:prec-rec} with the full proof provided in Appendix~\ref{app:theorem-1}.
\begin{theorem}
\label{the:prec-rec}
Denote $P(y^*=i) = p_i, i \in [Q]$. Assuming noise transition matrix $T$ and confusion matrix $C$ of a classifier, the expectations of \textit{precision} and \textit{recall} of the selected samples by Algorithm~\ref{alg:enkcvs} are then:
\end{theorem}
\begin{equation}
    \begin{aligned}
        Precision     &:= \frac{\sum\limits_{k=t}^{M} \tbinom{M}{k} \sum\limits_{i=1}^{Q} p_i T_{ii}^M C_{ii}^K (1 - C_{ii})^{M-k}}{\sum\limits_{k=t}^{M} \tbinom{M}{k} \sum\limits_{i=1}^{Q}\sum\limits_{j=1}^{Q} p_i T_{ij}^M C_{ij}^K (1 - C_{ij})^{M-k}}, \\
        Recall        &:= \frac{\sum\limits_{k=t}^{M} \tbinom{M}{k} \sum\limits_{i=1}^{Q} p_i T_{ii}^M C_{ii}^K (1 - C_{ii})^{M-k}}{\sum\limits_{i=1}^{Q} p_i T_{ii}}.
    \end{aligned}
\end{equation}
\begin{corollary}
\label{cor:symmetric}
For the special case where both the noise matrix and the confusion matrix are symmetric, with $C_{ii}=q$, $C_{ij}=\frac{1-q}{Q-1}$,  $M=1$, the \textit{precision} and \textit{recall} can be simplified as follows:
\end{corollary}
\begin{equation}
    \label{eq:prec-rec-sym}
    \begin{aligned}
        Precision     &:= \frac{(1-\epsilon)q}{(1-\epsilon)q + \epsilon (1-q) / (Q-1)}, \\ 
        Recall      &:= q.
    \end{aligned}
\end{equation}
% \begin{equation}
% \label{eq:prec-rec-sym}
%     \begin{aligned}
%         Precision     &:= \frac{(1-\epsilon)q}{(1-\epsilon)q + \epsilon (1-q) / (Q-1)},  
%         Recall        &:= q.
%     \end{aligned}
% \end{equation}
The performance is thus dependent on both the classifier accuracy $q$ and the noise transition matrix.
Furthermore, from Equation~\ref{eq:prec-rec-sym} we can see that the only way to improve both \textit{precision} and \textit{recall} is to improve $q$. Namely, to improve the accuracy of the classifier $f$ in Algorithm~\ref{alg:nkcvs}. 

The number of splits $K$ in Algorithm~\ref{alg:nkcvs} can be tuned for this purpose. In general, a higher value for $K$ will give better results since the training of classifier $f$ is augmented with more training samples, giving more accurate predictions and thus a higher $q$. 
For \ealname in Algorithm~\ref{alg:enkcvs}, we can further tune the number of iterations $M$ and the threshold $t$. This will not directly or consistently increase $q$, but will act as a regularizer to enhance the \textit{precision} at the cost of the \textit{recall}, or vice versa. 
An empirical study of the impact of $K$, $M$ and $t$ is found in Section~\ref{subsec:parameter-sensitivity}.
Although Corollary~\ref{cor:symmetric} is a special case for symmetric noise, we experimentally show in Section~\ref{sec:experiments} that for asymmetric and real-world noise, our algorithm achieves competitive results with state-of-the-art methods.

\subsection{Label re-weighting based on predicted labels}
\label{subsec:re-weight}
To make use of all available information, we do not simply discard the samples not selected by \ealname. Instead, we decrease the weight given to these samples during training of the final network. In Algorithm~\ref{alg:enkcvs} (\ealname), for each sample $(\mathbf{x}, y)$, we obtain $M$ predicted labels. We denote these as $\hat{y}^j, j=1,...,M$. We denote the label with the most occurrences as $\hat{y}$ and use it as a pseudo label. The distribution of the predicted labels is denoted as $\mathcal{P}$, and the entropy as $H(\mathcal{P})$, where $H(\mathcal{P}) = - \mathbb{E}_{\mathcal{P}} (\log \mathcal{P})$. 

We only re-weight samples $(\mathbf{x}, y) \notin \mathcal{SS}$. The loss using the original label $y$ and the pseudo label $\hat{y}$ are computed, and the weight between the two are determined by $\beta$ as follows:
\begin{equation}
    \beta \mathcal{L}(\mathbf{x}, y) + (1-\beta) \mathcal{L}(\mathbf{x}, \hat{y}).
\end{equation}
The weight $\beta$ is based on the label entropy and set to be:
\begin{equation}
    \beta = \frac{H(\mathcal{P})}{\log Q},
\end{equation}
where $\log Q$ is the maximum possible value of $H(\mathcal{P})$ and hence used to normalize $\beta$ to [0,1]. Thus, the weight given to $\mathcal{L}(\mathbf{x}, \hat{y})$ will decrease with increased uncertainty in $\hat{y}$.

By including the selected samples and a tuning parameter $\gamma$, we obtain the complete loss function:
\begin{align}
\label{eq:full-loss}
    \frac{1}{|\mathcal{SS}|} &\sum_{(\mathbf{x}, y) \in \mathcal{SS}} \mathcal{L}(\mathbf{x}, y) + \nonumber \\
     \frac{\gamma}{|\mathcal{D}| - |\mathcal{SS}|} & \sum_{(\mathbf{x}, y) \notin \mathcal{SS}} \big( \beta \mathcal{L}(\mathbf{x}, y) + (1-\beta) \mathcal{L}(\mathbf{x}, \hat{y}) \big),
\end{align}
where $\mathcal{L}$ is the standard cross-entropy loss.
%In our work, the loss function  is set to the standard cross-entropy loss.

\begin{comment}

\paragraph{Right samples:}
$H(\mathcal{P}) \to 0$ and $y = \hat{y}$.

% $\mathcal{RS} = \{(x,y) \in \mathcal{D}, y = \hat{y},  \sum\limits_{j=1}^{M} I(y = \hat{y}^j ) \geq t \}$
\paragraph{Wrong samples:}
$H(\mathcal{P}) \to 0$ and $y \neq \hat{y}$
\paragraph{Hard to tell samples:}
$H(\mathcal{P}) \to \log Q$

% $\mathcal{L}(\mathbf{x}, y) + (\alpha - \beta \frac{H(\mathcal{P})}{\log Q}) \mathcal{L}(\mathbf{x}, \hat{y}), \alpha \geq \beta$

Loss that is used:\\
% $\mathcal{US} = \{ i | i \notin \mathcal{SS}, i=1...n\}, \mathcal{Y}={y_i | i = 1...n}$,

% $loss=\sum\limits_{i \in \mathcal{SS}} \mathcal{L}(x_i, y_i) + 
% \gamma \sum\limits_{i \notin \mathcal{SS}} (\beta \mathcal{L}(x_i, y_i) +
%  (1-\beta) \mathcal{L}(x_i, \hat{y}_i))$, $ y_i \in \mathcal{Y},  \hat{y}_i \in \hat{\mathcal{Y}}
% $

$\mathop{\mathbb{E}}\limits_{(\mathbf{x}, y) \in \mathcal{SS}} (\mathcal{L}(\mathbf{x}, y)) + \gamma \mathop{\mathbb{E}}\limits_{(\mathbf{x}, y) \notin \mathcal{SS}} (\beta \mathcal{L}(\mathbf{x}, y) + (1-\beta) \mathcal{L}(\mathbf{x}, \hat{y})) 
$, 
where $\beta = \frac{H(\mathcal{P})}{\log Q}$.

\end{comment}

\section{Experiments}
\label{sec:experiments}
In this section, we demonstrate the validity and robustness of the proposed method when training on data with label noise. 
The section is divided into four parts. We begin by introducing the experimental setup. This is followed by a validation of the effectiveness of \ealname in identifying clean samples and a parameter analysis of how different parameter settings can affect the results. Finally, we show that our method is robust and can outperform state-of-the-art methods, both on datasets with artificially added noise and datasets with real-world noisy labels. 
%\blfootnote{Our source code is available online at \url{https://github.com/x/y}.}

\subsection{Experimental setup}
\label{subsec:exp-setup}
We extensively validate our method on multiple benchmark datasets, namely MNIST~\cite{lecun1998gradient}, CIFAR-10 and CIFAR-100 \cite{krizhevsky2009learning}, TREC~\cite{li2002learning}, Clothing-1M~\cite{xiao2015learning}, and WebVision~\cite{li2017webvision}.
We use symmetric and asymmetric noise as defined in Section~\ref{sec:preliminaries} to manually corrupt the labels in the training data $\mathcal{D}_{train}$ with different noise ratios. The labels in the testing data $\mathcal{D}_{test}$ are kept clean. For Clothing-1M and WebVision, we do not introduce any artificial noise since the datasets are naturally noisy. 
To obtain the final test accuracy, the DNN is retrained using the selected samples $\mathcal{SS}$ and the re-weighted non-selected samples. The test evaluation is done with $\mathcal{D}_{test}$.

For the real-world datasets, we follow previous works~\cite{li2020dividemix,chen2019understanding} and use ResNet-50 with weights pre-trained on ImageNet for Clothing-1M and inception-resnet v2~\cite{szegedy2016inception} for WebVision.
We discard aby labeled training images provided in the datasets.
%Any labeled training images provided in the datasets have been discarded and are not used.
For WebVision, we use the first 50 classes of the Google image subset and evaluate the results on the provided validation dataset. A summary of the datasets and the full details on the experimental setup are provided in Appendix~\ref{app:dataset-summary-and-exp-setup}.

The default parameters of \ealname are set as follows, $K=10$, $M=5$ and $t=2$. For each iteration over $M$, $D_{train}$ is split into random folds and the network is randomly initialized.

\subsection{Method effectiveness and parameter sensitivity}
\label{subsec:parameter-sensitivity}

\begin{table}[t]
	\begin{center}
		\begin{tabular}{cccccc}
			\toprule
			$\epsilon$ & $|\mathcal{CS}|$ & $|\mathcal{SS}|$ & $|\mathcal{CSS}|$ & $Precision$ & $Recall$ \\
			\midrule
			0.0    & 50000 & 46659 & 46659 & 100.0\%  & 93.32\% \\
			0.2    & 40000 & 37111 & 36958 & 99.59\%  & 92.40\% \\
			0.4    & 30000 & 27505 & 27130 & 98.64\%  & 91.68\% \\
% 			0.5    & 25162 & 24343 & 22967 & 94.35\%  & 91.28\% \\
			0.6    & 20000 & 18127 & 17304 & 95.46\%  & 86.52\% \\
			0.8    & 10000 & 9070  & 6602  & 72.79\%  & 66.02\% \\
			\bottomrule
		\end{tabular}
	\end{center}
	\caption{The performance of \ealname on CIFAR-10 with different noise ratios $\epsilon$. The parameters $K, M$ and $t$ are set to $10$, $5$ and $2$, respectively.}
	\label{tab:diff_noise}
\end{table}

\begin{comment}
\begin{figure}
    \centering
    \subfigure[Varying $K$ with $M=4$\newline and $t=2$. ]{\label{fig:K_prec_recall}\includegraphics[width=45mm]{fig/old_K_prec_recall.pdf}}
    \subfigure[Varying $K$ with $M=4$\newline and $t=2$. ]{\label{fig:K_prec_recall}\includegraphics[width=45mm]{fig/old_K_prec_recall.pdf}}
    \subfigure[Varying $M$ with $K=10$ and $t=M-1$. ]{\label{fig:M_prec_recall}\includegraphics[width=45mm]{fig/old_M_prec_recall.pdf}}
    \caption{The sensitivity of \textit{precision} and \textit{recall} when adjusting $K$ and $M$. Experiments done on $D_{train}$ from CIFAR-10 with label noise $\epsilon=0.2$ and $\epsilon=0.8$.}
\end{figure}
\end{comment}

% Merged figure
\begin{figure*}[t]
    \centering
    \includegraphics[trim={2cm 0.25cm 0.25cm 0.9cm}, clip, width=\textwidth]{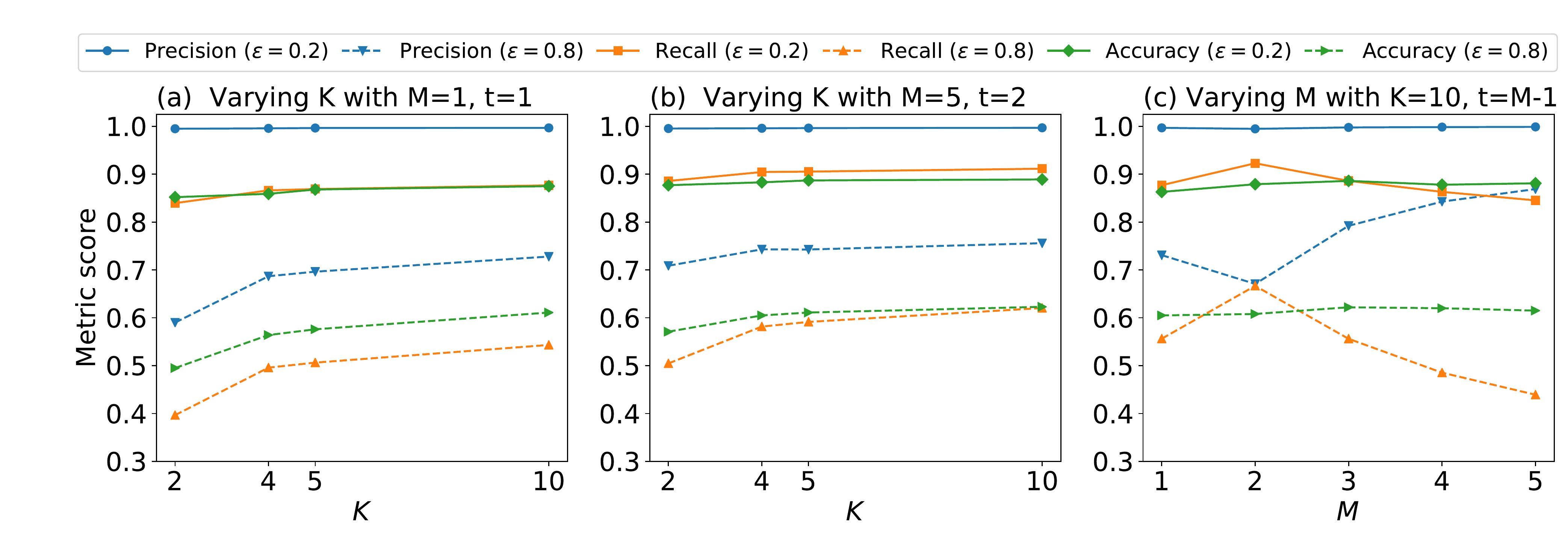}
    \caption{The sensitivity of \textit{precision} and \textit{recall} when adjusting $K$ and $M$ using $D_{train}$ from CIFAR-10.
    The accuracy is measured on $D_{test}$ using the model trained on the selected samples $\mathcal{SS}$.
    }
    \label{fig:prec_recall_acc}
\end{figure*}

We conduct experiments on how effective \ealname is in identifying the clean samples. 
In particular, we verify our claim that increasing the number of splits $K$ will have a significant positive impact on both \textit{precision} and \textit{recall}, as defined in Equation~\ref{eq:prec-rec}. Moreover, the dynamics between $M$ and $t$ are investigated in detail. Here, we do not re-weight the non-selected samples in order to examine the pure effect of the parameters on the selected samples. For this purpose, we use the training data set $D_{train}$ from CIFAR-10 with varying degrees of label noise and empirically show how the different parameters of \ealname affect the result.

The parameter that has the largest influence on the results is without a doubt the noise ratio $\epsilon$. Table~\ref{tab:diff_noise} shows the \textit{precision} and \textit{recall} as well as the sizes of the different sample sets in the presence of different noise ratios.
Predictably, the accuracy of the sample selection drops with an increase in label noise, i.e., the baseline with $\epsilon=0.0$ achieves the best result.
There is a significant drop in both \textit{precision} and \textit{recall} when increasing $\epsilon$ from $0.6$ to $0.8$ showing that obtaining a good model in the case of severe label noise is particularly challenging.

\paragraph{Impact of increasing $\mathbf{K}$:}
In Figure~\ref{fig:prec_recall_acc}(b) we see the results of \ealname for varying values of $K$ while the plain \alname (i.e., $M=1$) results are shown in Figure~\ref{fig:prec_recall_acc}(a).
As clearly illustrated in the figures, the generalization performance steadily increases with higher $K$.
Moreover, the increase is more pronounced at higher noise ratios.
The increase in both \textit{precision} and \textit{recall} with higher $K$ indicates that more of the clean samples in $\mathcal{CS}$ are found while the purity of the selected samples $\mathcal{SS}$ is increased. This improvement of $\mathcal{SS}$ is reflected in the final accuracy which is also higher with larger $K$.
However, the expected performance gain of increasing $K$ will have diminishing returns, and any performance gain needs to be weighed against the increased computation cost. 
%Note that $K=2$ is the lowest possible value since a sub-partition of the data is used for training and the other part is for selecting clean samples. On the other hand, the upper bound of $K$ is $|D_{train}|-1$, i.e., leave-one-out cross-validation.

\paragraph{Impact of increasing $\mathbf{M}$:}
We investigate the impact of increasing $M$ while fixing $t=M-1$ (for $M=1$ we use $t=1$), the results are in Figure~\ref{fig:prec_recall_acc}(c). $M$ will act as a regularizer, namely, a higher $M$ will increase in \textit{precision} and decrease \textit{recall}. This indicates that the selected sample set $\mathcal{SS}$ contains fewer samples but has a higher proportion of clean samples.
The resulting accuracy on the model trained on $\mathcal{SS}$ generally increases slightly with higher $M$, emphasizing the relative importance of a higher \textit{precision}. 

In Figure~\ref{fig:prec_recall_acc}(a) and Figure~\ref{fig:prec_recall_acc}(b) the the difference between plain \alname (i.e., setting $M=1$) and \ealname with $M=5$ is illustrated.
\ealname outperforms \alname, especially for lower values of $K$, demonstrating the significance of $M$.
The outcome of fixing $t$ while increasing $M$ can be observed in Figure~\ref{fig:prec_recall_acc}(c), 
where $M$ changes from 1 to 2 while $t$ is kept at 1. This will have the opposite effect to increasing $M$ while keeping $t$ static, i.e., $\mathcal{SS}$ will have more but noisier samples.
The behavior of $M$ and $t$ is thus very flexible and can be adjusted depending on the scenario.
In addition, the label re-weighting scheme will also benefit from larger $M$.

% Moved here to make sure it's on an earlier page.
\begin{table*}[t]
    % \scriptsize
    % \footnotesize
	\begin{center}
        \begin{tabular}{cc|ccccc|cc}
        \hline
        \multirow{2}*{Datasets} & \multirow{2}*{Methods} & \multicolumn{5}{c}{Symmetric Noise} & \multicolumn{2}{|c}{Asymmetric Noise}\\
        \cline{3-9}
        & & 0.0 & 0.2 & 0.4 & 0.6 & 0.8 & 0.2 & 0.4\\
        \cline{1-9}
        \multirow{4}*{MNIST}
        & CE          & \textbf{99.3 $\pm$ 0.1} & 98.6 $\pm$ 0.1 & 98.1 $\pm$ 0.2 & 97.0 $\pm$ 0.2 & 81.5 $\pm$ 0.5 & 93.1 $\pm$ 0.1& 81.1 $\pm$ 0.5\\
        % & Forward     & 99.3 $\pm$ 0.1 & 98.5 $\pm$ 0.1 & 98.0 $\pm$ 0.2 & 96.8 $\pm$ 0.1 & 81.3 $\pm$ 0.2 & 93.1 $\pm$ 0.1& 81.1 $\pm$ 0.1\\
        % & LSR         & 99.3 $\pm$ 0.1 & 98.6 $\pm$ 0.1 & 98.0 $\pm$ 0.1 & 96.7 $\pm$ 0.2 & 82.4 $\pm$ 0.5 & 93.1 $\pm$ 0.1& 81.1 $\pm$ 0.1\\
        % & GCE         & \textbf{99.3 $\pm$ 0.1} & 99.1 $\pm$ 0.1 & 98.5 $\pm$ 0.1 & 98.0 $\pm$ 0.2 & 81.5 $\pm$ 0.5 & 96.7 $\pm$ 0.2& 81.3 $\pm$ 0.4\\
        & Co-teaching & 99.2 $\pm$ 0.1 & 99.2 $\pm$ 0.1 & \textbf{99.1 $\pm$ 0.1} & 98.4 $\pm$ 0.1 & 88.2 $\pm$ 0.5 & 97.1 $\pm$ 0.2& 88.8 $\pm$ 0.5\\
        & SL          & \textbf{99.3 $\pm$ 0.1} & 99.2 $\pm$ 0.1 & 99.0 $\pm$ 0.1 & 98.3 $\pm$ 0.1 & 91.4 $\pm$ 0.1 & \textbf{99.1 $\pm$ 0.1}& 98.0 $\pm$ 0.1\\
        & E-NKCVS        & \textbf{99.3 $\pm$ 0.1} & \textbf{99.3 $\pm$ 0.1} & \textbf{99.1 $\pm$ 0.1} & \textbf{98.5 $\pm$ 0.1} & \textbf{91.9 $\pm$ 0.1} & \textbf{99.1 $\pm$ 0.2}& \textbf{98.4 $\pm$ 0.2}\\
        \hline
        \multirow{4}*{CIFAR-10}
        & CE          & \textbf{89.7 $\pm$ 0.1} & 83.5 $\pm$ 0.1 & 78.8 $\pm$ 0.2 & 69.9 $\pm$ 0.6 & 41.5 $\pm$ 0.5 & 85.9 $\pm$ 0.2& 78.5 $\pm$ 0.6\\
        % & Forward     & 88.7 $\pm$ 0.1 & 83.5 $\pm$ 0.2 & 78.7 $\pm$ 0.1 & 70.0 $\pm$ 0.5 & 41.9 $\pm$ 0.5 & 93.1 $\pm$ 0.1& 81.1 $\pm$ 0.1\\
        % & LSR         & 88.2 $\pm$ 0.1 & 83.7 $\pm$ 0.2 & 78.3 $\pm$ 0.2 & 67.8 $\pm$ 0.8 & 42.2 $\pm$ 0.4 & 93.1 $\pm$ 0.1& 81.1 $\pm$ 0.1\\
        % & GCE         & 86.5 $\pm$ 0.1 & 84.5 $\pm$ 0.2 & 81.9 $\pm$ 0.3 & 76.3 $\pm$ 0.3 & 44.7 $\pm$ 0.6 & 84.6 $\pm$ 0.3& 75.3 $\pm$ 0.5\\
        & Co-teaching & 89.4 $\pm$ 0.2 & 86.6 $\pm$ 0.3 & 84.1 $\pm$ 0.8 & 81.1 $\pm$ 0.6 & 22.5 $\pm$ 3.6 & 86.8 $\pm$ 0.4& 75.5 $\pm$ 0.5\\
        & SL          & 89.5 $\pm$ 0.1 & 87.6 $\pm$ 0.1 & 85.3 $\pm$ 0.1 & 80.1 $\pm$ 0.1 & 59.5 $\pm$ 0.5 & 88.2 $\pm$ 0.1& 80.6 $\pm$ 0.4\\
        & E-NKCVS        & \textbf{89.7 $\pm$ 0.1} & \textbf{89.0 $\pm$ 0.1} & \textbf{86.3 $\pm$ 0.2} & \textbf{83.1 $\pm$ 0.2} & \textbf{63.5 $\pm$ 0.4} & \textbf{88.9 $\pm$ 0.2}& \textbf{85.1 $\pm$ 0.3} \\
        \hline
        \multirow{4}*{CIFAR-100}
        & CE          & \textbf{69.1 $\pm$ 0.6} & 61.1 $\pm$ 0.4 & 51.4 $\pm$ 0.6 & 27.6 $\pm$ 1.2 & 7.7  $\pm$ 1.5 & 63.0 $\pm$ 0.3 & 61.8 $\pm$ 0.4\\
        % & Forward     & 66.3 $\pm$ 0.1 & 59.4 $\pm$ 0.1 & 52.2 $\pm$ 0.2 & 32.1 $\pm$ 0.1 & 7.2  $\pm$ 0.2 & 93.1 $\pm$ 0.1& 81.1 $\pm$ 0.1\\
        % & LSR         & 65.6 $\pm$ 0.1 & 60.6 $\pm$ 0.1 & 51.0 $\pm$ 0.1 & 24.7 $\pm$ 0.2 & 9.2  $\pm$ 0.5 & 93.1 $\pm$ 0.1& 81.1 $\pm$ 0.1\\
        % & GCE         & 64.4 $\pm$ 0.1 & 59.1 $\pm$ 0.1 & 53.2 $\pm$ 0.1 & 36.2 $\pm$ 0.2 & 8.5  $\pm$ 0.5 & 63.1 $\pm$ 0.2 & 61.3 $\pm$ 0.3\\
        & Co-teaching & 66.2 $\pm$ 0.5 & 61.3 $\pm$ 0.5 & 52.3 $\pm$ 0.8 & 41.1 $\pm$ 1.6 & 5.5 $\pm$ 2.6 & 63.2 $\pm$ 0.3 & 62.2 $\pm$ 0.5\\
        & SL          & 68.2 $\pm$ 0.1 & 62.1 $\pm$ 0.1 & 55.3 $\pm$ 0.1 & 43.4 $\pm$ 0.1 & 15.5 $\pm$ 0.1 & 66.1 $\pm$ 0.2 & 63.1 $\pm$ 0.4\\
        & E-NKCVS        & \textbf{69.1 $\pm$ 0.5} & \textbf{64.8 $\pm$ 0.2} & \textbf{59.6 $\pm$ 0.4} & \textbf{47.9 $\pm$ 0.3} & \textbf{26.3 $\pm$ 0.3} & \textbf{67.9 $\pm$ 0.3} & \textbf{64.5 $\pm$ 0.5}\\
        \hline
        
        \multirow{4}*{TREC}
        & CE          & \textbf{96.5 $\pm$ 0.5} & 93.5 $\pm$ 0.8 & 90.1 $\pm$ 1.5 & 77.6 $\pm$ 4.2 & 26.3 $\pm$ 7.5 & 93.0 $\pm$ 1.0& 79.3 $\pm$ 4.5\\
        % & LSR          & 96.5 $\pm$ 0.5 & 93.4 $\pm$ 0.2 & 90.1 $\pm$ 1.5 & 77.6 $\pm$ 4.2 & 26.3 $\pm$ 7.5 & 93.2 $\pm$ 0.5& 93.2 $\pm$ 0.5\\
        % & GCE         & 95.6 $\pm$ 0.5 & 93.9 $\pm$ 0.4 & 89.4 $\pm$ 1.9 & 75.8 $\pm$ 5.3 & 24.6 $\pm$ 6.8 & 93.8 $\pm$ 0.5& 81.2 $\pm$ 5.5\\
        & Co-teaching & 95.2 $\pm$ 0.5 & 92.3 $\pm$ 0.8 & 90.3 $\pm$ 0.8 & 81.3 $\pm$ 1.6 & 25.5 $\pm$ 4.6 & 92.8 $\pm$ 0.6& 77.3 $\pm$ 3.8\\
        & SL          & 96.4 $\pm$ 0.4 & 93.7 $\pm$ 0.5 & 92.2 $\pm$ 0.6 & 83.5 $\pm$ 2.2 & 30.2 $\pm$ 4.9 & 94.0 $\pm$ 0.5 & 83.3 $\pm$ 4.2\\
        & E-NKCVS       & 96.4 $\pm$ 0.2 & \textbf{95.0 $\pm$ 0.5} & \textbf{93.8 $\pm$ 0.5} & \textbf{89.9 $\pm$ 0.6} & \textbf{35.5 $\pm$ 2.5} & \textbf{95.0 $\pm$ 0.6}& \textbf{88.0 $\pm$ 1.2}\\
        \hline
        \end{tabular}
	\end{center}
		\caption{Test accuracy (\%, average over 5 runs) with different label noise ratios $\epsilon$. We train on the manually corrupted $D_{train}$ and test on the clean $D_{test}$ data. The best results are highlighted in bold.}
	\label{exp:result}
\end{table*}

\subsection{Comparison to the state-of-the-art}
\label{subsec:stateart}

We compare our algorithm with the standard cross-entropy loss and multiple state-of-the-art methods made for dealing with noisy labels. 

\begin{itemize}
    \item CE: Basic cross-entropy loss.
    % \item Forward~\cite{patrini2017forward}: Using loss correction during training by multiplying the prediction with the ground-truth noise transition matrix $T$.
    % \item LSR~\cite{pereyra2017lsr}: Adding a label smoothing regularization term to the loss that penalizes low entropy output distributions.
    %\item GCE~\cite{zhang2018gce}: Using a noise-robust loss based on the negative Box-Cox transformation. %, that generalizes both MAE and CE.
    \item Co-teaching~\cite{han2018coteach}: Simultaneously training two networks and let them teach each other. Samples selected by one network in a mini-batch are used for back-propagation in the other.
    \item SL~\cite{wang2019sl}: A symmetric cross-entropy learning approach that tries to balance between sufficient learning and robustness to noisy labels.
    %\item MLNT~\cite{Li2019LearningTL}: Optimize a meta-objective that aims to find noise-tolerant model parameters together with a student-teacher neural network.
    %\item CleanNet~\cite{lee2017cleannet}: A transfer learning-based joint neural embedding network. However, it requires some auxiliary cleaned labeled data during training.
    
    \item DMI~\cite{Xu2019L_DMIAI}: Uses a determinant-based mutual information robust loss to train the DNNs.
    %Using a loss function based on Determinant based Mutual Information (DMI) which is a generalized version of mutual information.
    
    % https://arxiv.org/pdf/1909.03388.pdf
    
    % Xu et al. (2019) introduce a Determinantbased Mutual Information (DMI) loss for robust fine-tuning of a CE pre-trained model. 
    
    % [17] propose the Determinant-based Mutual Information, a generalized version of mutual information that is provably insensitive to noise patterns and amounts.

    % which uses mutual information based robust loss to train the DNNs.

    %\item MetaCleaner~\cite{Zhang_2019_CVPR}: Estimates confidence scores for a subset of samples labeled as the same class and hallucinates a clean representation based on the scores, to be used in training the classifier.
    
    \item MentorNet~\cite{jiang2017mentornet}: Trains a teacher network to teach a student network by providing a sample weighting scheme.

    \item MentorMix~\cite{jiang2020beyond}: Proposes a new robust loss by mixing curriculum learning from MentorNet~\cite{jiang2017mentornet} and vicinal risk minimization.
    % http://proceedings.mlr.press/v119/jiang20c/jiang20c.pdf
    
    \item DivideMix~\cite{li2020dividemix}: Dynamically divides the data based on label confidence and trains two networks in a semi-supervised manner based on MixMatch.
    
    % https://arxiv.org/pdf/2002.07394.pdf
    
    %Meanwhile, DivideMix [91] is an extension of the semi-supervised data augmentation technique called MixMatch [105]. A two-component and one-dimensional Gaussian mixture model is fitted to the training loss to obtain the confidence of an annotated label. By setting a confidence threshold, the training data is categorized into a labeled set and an unlabeled set. Subsequently, MixMatch is employed to train a DNN for the transformed data.
    
    %In a similar spirit, DivideMix [22] uses two networks to perform sample selection via a two-component mixture model, and applies the semi-supervised learning technique MixMatch [4].

    %In particular, DivideMix models the per-sample loss distribution with a mixture model to dynamically divide the training data into a labeled set with clean samples and an unlabeled set with noisy samples, and trains the model on both the labeled and unlabeled data in a semi-supervised manner. To avoid confirmation bias, we simultaneously train two diverged networks where each network uses the dataset division from the other network. During the semi-supervised training phase, we improve the MixMatch strategy by performing label co-refinement and label co-guessing on labeled and unlabeled samples, respectively.

\end{itemize}

\noindent
The baselines originally evaluate on different datasets and in the following evaluation we keep these distinctions. Accuracy on real-world datasets is reported as in the original papers while those baselines that utilize synthetic data are rerun for a fair comparison.

The experimental results on the datasets with artificial noise are summarized in Table~\ref{exp:result} with \ealname using the default parameters as presented in Section~\ref{subsec:exp-setup}. Our solution outperforms the baseline methods and achieves the best test accuracy at all levels and types of label noise.
The advantage of our method is more significant when the noise ratio is more severe and the difficulty increases. This is, in particular, discernible on CIFAR-100 which is a more challenging dataset.

In the scenario with no corrupted labels (i.e., $\epsilon=0$), CE is the best baseline since it focuses on fitting the data instead of dealing with noisy labels. For the other methods, the adjustments made to the loss or the network architecture to account for label noise are at best wasted, and in many cases detrimental. In contrast, our \ealname algorithm will have close to identical test accuracy as compared to CE, implying that our method has a minimal negative impact when there is no or little label noise present.

\subsection{Experiments on real-world noisy datasets}
\label{subsec:experiment_real_world}
We further assess the capabilities of \ealname and its practical usage on two datasets with real-world noisy labels, Clothing-1M and (mini) WebVision. The results can be seen in Table~\ref{tab:real-world}.

As shown, \ealname achieves an accuracy of 75.0\% on Clothing-1M, improving the prior state-of-the-art without the use of the auxiliary training labels. Running without any additional considerations for the label noise, i.e., using a basic CE loss, a test accuracy of 69.0\% is obtained. Thus, to be conscious of and take steps to rectify mislabeled samples can give substantial model improvements on real-world datasets.

For WebVision, the test accuracy of \ealname is competitive to recent published works and improves slightly upon the prior state-of-the-art. The results imply that our relatively simple \ealname method is reliable and effective on datasets containing real-world noisy labels. Here, the CE baseline is surprisingly strong with a test accuracy of 74.0\%, only outperformed by our method and the two best baseline methods.  WebVision has a relatively lower noise level of around 20\%~\cite{li2017webvision} compared to Clothing-1M with close to 40\% label noise~\cite{xiao2015learning} which could explain part of this discrepancy.

% Learning from Noisy Labels with Complementary Loss Functions
%Our method falls short of the state-of-the-art 74.76. We think this limitation is because the Clothing1M dataset contains instance-dependent noisy labels, and MAE and APL are not robust to this type of noise.

\begin{table}[t]
	\begin{center}
		\begin{tabular}{lcc}
			\toprule
			 Method & Clothing-1M & WebVision \\
			 \midrule
			CE & 69.0 & 74.0 \\
			MentorNet & -  & 63.0 \\
			SL & 71.0 & - \\
			DMI & 72.5 & - \\
			Co-teaching & - & 63.6 \\
			MentorMix & 74.3 & 76.0 \\
			DivideMix & 74.8 & 77.3 \\
			E-NKCVS & \textbf{75.0} & \textbf{77.6} \\
			\bottomrule
		\end{tabular}
	\end{center}
	\caption{Comparison with state-of-the-art methods in test accuracy (\%) on Clothing1M and (mini) WebVision. Results for baselines are copied from the original papers or, if missing, from Li et al.~\protect\shortcite{li2020dividemix}.}
	\label{tab:real-world}
\end{table}

% Original
%CE & 69.0 & 74.0 \\
%MentorNet & -  & 63.0 \\
%SL & 71.02 & - \\
%DMI & 72.46 & - \\
%Co-teaching & - & 63.58 \\
%MentorMix & 74.3 & 76.0 \\
%DivideMix & 74.76 & 77.32 \\
%E-NKCVS & \textbf{74.99} & \textbf{77.56} \\
\section{Conclusion}
In this paper, we propose Ensemble Noise-Robust $K$-fold Cross-Validation Selection (\ealname) to deal with the noisy label problem by selecting likely clean samples to use for model training. For non-selected samples, we further propose to use an entropy-based label re-weighting scheme based on the given label and the predicted labels. 
The effectiveness of our solution is verified on multiple datasets that are manually corrupted with different levels of symmetric and asymmetric label noise. We show that \ealname consistently outperforms existing methods at all levels of label noise. Particularly on the more complex and challenging dataset, CIFAR-100, we achieve a significant improvement over the second-best approach at high noise ratios. 
Experiments on two large real-world datasets with natural label noise, Clothing-1M and WebVision, further emphasize the usefulness of our method, and we show that our method achieves state-of-the-art test accuracy on both datasets. 
An extensive empirical hyperparameter analysis is provided that demonstrates the versatility of our proposed method. 
Moreover, due to the method's relative simplicity, it can easily be incorporated into existing DNN algorithm architectures to enhance their robustness against label noise.

\bibliographystyle{named}
{\small \bibliography{main.bib}}  % can add {\small ...} here if needed (allowed, see: https://ijcai-21.org/ijcai-21-faq/)
%\nobibliography{main.bib} % To not show reference table

% This need to be considered separately when creating the final pdf, see line 80 here above.
\appendix
% Can have some short intro text here

\newpage
\appendix

% We only have one extra proof now so removing the subsection.
%\section{Proofs and Extended Theoretical Analysis}

\section{Proof of Theorem 1}
\label{app:theorem-1}

\setcounter{theorem}{0}
% \begin{theorem}
% Denote $P(y^*=i) = p_i, i \in [Q]$. Assuming a balanced dataset with a label noise transition matrix $T$ and confusion matrix $C$ of a classifier, the expectations of \textit{precision} and \textit{recall} of the selected samples by Algorithm~\ref{alg:enkcvs} are then:
% \end{theorem}

% \begin{equation}
%     \begin{aligned}
%         Prec     &:= \frac{\sum\limits_{k=t}^{M} C_M^k \sum\limits_{i=1}^{Q} p_i T_{ii}^M C_{ii}^K (1 - C_{ii})^{M-k}}{\sum\limits_{k=t}^{M} C_M^k \sum\limits_{i=1}^{Q}\sum\limits_{j=1}^{Q} p_i T_{ij}^M C_{ij}^K (1 - C_{ij})^{M-k}} , 
%         Rec        &:= \frac{\sum\limits_{k=t}^{M} C_M^k \sum\limits_{i=1}^{Q} p_i T_{ii}^M C_{ii}^K (1 - C_{ii})^{M-k}}{\sum\limits_{i=1}^{Q} p_i T_{ii}}.
%     \end{aligned}
% \end{equation}

\begin{proof}
\begin{equation*} 
    \begin{aligned}
        P(x \in \mathcal{CS} )   &= \sum_{i=1}^{Q} p_i T_{ii}, \\
        P(x \in \mathcal{SS} )   &= \sum_{k=t}^{M} C_M^k \sum_{i=1}^{Q}\sum_{j=1}^{Q} p_i T_{ij}^M C_{ij}^K (1 - C_{ij})^{M-k} \\
        P(x \in \mathcal{CSS})   &= \sum_{k=t}^{M} C_M^k \sum_{i=1}^{Q} p_i T_{ii}^M C_{ii}^K (1 - C_{ii})^{M-k}  \\
    \end{aligned}
\end{equation*}
Inserting the above into the definitions of \textit{precision} and \textit{recall} in Equation~\ref{eq:prec-rec}, we obtain the desired results.
\end{proof}

\section{Dataset Summary and Experimental Setup}
\label{app:dataset-summary-and-exp-setup}

\subsection{Dataset Summary}
\label{app:dataset-summary}
The datasets used in the experimental part of the paper are introduced one by one below. A brief overview can be found in Table~\ref{tab:dataset}.

\vspace{.5\baselineskip}\noindent
\textbf{MINST}~\cite{lecun1998gradient} is a popular but small dataset of handwritten digits. This is a balanced dataset with 10 classes, each of which has 6,000 training and 1,000 testing images.

\vspace{.5\baselineskip}\noindent
\textbf{CIFAR-10}~\cite{krizhevsky2009learning} contains images with human-annotated labels. There are 10 classes, airplane, automobile, bird, cat, deer, dog, frog, horse, ship, and truck. The dataset is balanced, with 5,000 images in the training set and 1,000 in the test set for each class.

\vspace{.5\baselineskip}\noindent
\textbf{CIFAR-100}~\cite{krizhevsky2009learning} is similar to CIFAR-10 but contains 100 classes instead of 10. This dataset is also balanced, with 500 training images and 100 testing images per class. This dataset is more complex as compared to CIFAR-10 with both more classes as well as fewer images per class. 

\vspace{.5\baselineskip}\noindent
\textbf{TREC}~\cite{li2002learning} (Text REtrieval Conference Question Classification) is a text dataset for question classification. The idea behind the dataset is that if a question can be classified into a semantic class correctly, this will put constraints on potential answers.
The dataset has 6 labels which can be further separated into 50 second-level labels. In this work, we only use the 6 main labels. These are abbreviation, entity, description, human, location, and numeric. The average question sentence length is 10 and the vocabulary size is 8,700.

\vspace{.5\baselineskip}\noindent
\textbf{Clothing-1M}~\cite{xiao2015learning} is a real-world dataset containing 1 million images of clothing articles. There are 14 different classes: t-shirt, shirt, knitwear, chiffon, sweater, hoodie, windbreaker, jacket, down coat, suit, shawl, dress, vest, and underwear. The images have all been obtained from online shopping websites and the labels are created from the attached text. 
The accuracy of the labels is about 61.54\%. Some classes are more often confused with each other (e.g., sweater and knitwear), indicating that the dataset may contain both symmetric and asymmetric noise. 
There are 50,000 manually cleaned images provided for training. The validation and test sets contain 14,000 and 10,000 clean images, respectively.
The cleaned training and validation images are not used and only the 10,000 testing samples are used for evaluation.

%\textbf{Food-101N}~\cite{lee2017cleannet} is a dataset containing images of food. It shares the same classes as the Food-101~\cite{bossard14} dataset, however, it contains more images and has more label noise. The dataset has a total of $310{,}009$ images divided into 101 classes which are collected from Google, Bing, Yelp, and Tripadvisor. The estimated label accuracy is 80\%. The test accuracy is directly evaluated on the test set from Food-101.

\vspace{.5\baselineskip}\noindent
\textbf{WebVision}~\cite{li2017webvision} is a large-scale dataset with real-world noisy labels containing images. The whole dataset contains 2.4 million images collected from the web using the same classes as ImageNet~\cite{deng2009imagenet}. The noise level in WebVision is estimated to be around 20\%~\cite{li2017webvision}.
We follow previous works~\cite{chen2019understanding,li2020dividemix} and use the first 50 classes of the Google image subset. We evaluate the test accuracy on the provided validation dataset.

\begin{table}[t]
    \small
	\begin{center}
		\begin{tabular}{lcccc}
			\toprule
			Dataset & $|\mathcal{D}_{train}|$ & $|\mathcal{D}_{test}|$  & Q & Image size \\
			\midrule
			MNIST       & 60,000     & 10,000 & 10  & $28 \times 28 $  \\
			CIFAR-10    & 50,000     & 10,000 & 10  & $32 \times 32 \times 3$ \\
			CIFAR-100   & 50,000     & 10,000 & 100 & $32 \times 32 \times 3$ \\
			TREC        & 5,500       & 500    & 6   & - \\
			Clothing-1M & 1,000,000  & 10,526 & 14  & $224 \times 224 \times 3$ \\
			%Food-101N   & 310,009    & 25,250 & 101 & $224 \times 224 \times 3$ \\
			WebVision (mini)  & 69,544    & 2,500 & 50 & $299 \times 299 \times 3$ \\
			\bottomrule
		\end{tabular}
	\end{center}
	\caption{Dataset summary.}
	\label{tab:dataset}
\end{table}

\subsection{Experimental Setup}
\label{app:experimental-setup}

During training, in each iteration of $K$ in Algorithm~\ref{alg:nkcvs}, we use $10\%$ of the data in $D_*$ to be a validation set $\mathcal{D}_{val}$. Then, the optimal model parameter $\theta^*$ is obtained by
\begin{equation} \label{eq:para_val}
     \theta^{*} = \mathop{\arg\max}\limits_{\theta} \mathcal{M}(f(\mathbf{x}, \theta), y), (\mathbf{x}, y) \in \mathcal{D}_{val},
\end{equation}
where $f$ is trained on $\mathcal{D}_{*} \setminus \mathcal{D}_{val}$, and $\mathcal{M}(\cdot)$ is a metric function. Here, we set $\mathcal{M}$ to be the accuracy. Note that $\mathcal{D}_{val}$ will be different in each iteration since $\mathcal{D}_*$ will change.

\paragraph{Asymmetric noise:}
For the experiments using asymmetric noise, it requires pairs of labels to be flipped for some fraction of samples. These flips are done on pairs of similar classes in each dataset.
Following the setting of asymmetric noise in~\cite{wang2019sl}, in the MINST dataset, we flip $2 \rightarrow 7$, $3 \rightarrow 8$, $5 \leftrightarrow 6$, and $7 \rightarrow 1$.
For CIFAR-10, we flip TRUCK $\rightarrow$ AUTOMOBILE, BIRD $\rightarrow$ AIRPLANE, DEER $\rightarrow$ HORSE, and CAT $\leftrightarrow$ DOG.
In CIFAR-100, there are 20 super-classes each of which has 5 sub-classes.
To generate asymmetric noise, we randomly flip the labels of two sub-classes within each super-class.
For the TREC dataset, we flip abbreviation $\leftrightarrow$ entity, description $\leftrightarrow$ human, and location $\rightarrow$ human.\\

% Experimental setup
\noindent
We set the loss tuning parameter $\gamma$ in Equation~\ref{eq:full-loss} to $0.2$.
For the mixup data augmentation, $\alpha$ in Equation~\ref{eq:mixup} is set to $0.3$.
The network architectures and all further dataset-specific parameters are given as follows. 

\vspace{.5\baselineskip}\noindent
\textbf{MINST:} A simple 4-layer CNN network (two convolutional and two fully connected layers) is used. We train the network with stochastic gradient descent (SGD) with a momentum $0.9$. The learning rate is initially set to $0.01$ with a weight decay of $10^{-4}$. 
The training is run for 50 epochs and the learning rate is divided by 10 after 10 and 30 epochs.

\vspace{.5\baselineskip}\noindent
\textbf{CIFAR-10:} We use an 8-layer CNN with six convolutional and two fully connected layers. We train the network with SGD with a momentum $0.9$. Similar to MINST, we set the initial learning rate to $0.01$ with a weight decay of $10^{-4}$. We divide the learning rate by 10 after 40 and 80 epochs and run for a total of 120 epochs.

\vspace{.5\baselineskip}\noindent
\textbf{CIFAR-100:}
Due to the relatively larger and more complex dataset, we use a larger network, ResNet-44 \cite{he2016deep}. We train the network with SGD with a momentum $0.9$. The initial learning rate is set to $0.1$ and the weight decay to $5\cdot 10^{-3}$. The training is run for 150 epochs and we divide the learning rate by 10 after 80 and 120 epochs.

\vspace{.5\baselineskip}\noindent
\textbf{TREC:}
Since TREC is a text-based dataset, we use the pre-trained BERT-Base as our network. We use the Adam optimization algorithm with a learning rate of $6.5 \cdot 10^{-5}$ and run for 5 epochs with a batch size of 200. Note that no mixup data augmentation is used for this dataset.

\vspace{.5\baselineskip}\noindent
\textbf{Clothing-1M:}
Following~\cite{xiao2015learning,wang2019sl}, we use ResNet-50 with ImageNet pre-trained weights. The manually cleaned and labeled training images provided are not used and discarded. Evaluation is done on the provided clean testing set of $10{,}527$ images. For preprocessing, we resize the images to $256 \times 256$ and subtract the mean for each pixel. The images are then cropped at the center to a size of $224 \times 224$. We train the network with SGD for a single epoch with a learning rate of $0.001$ and a batch size of 200. We set $M=10$, $K=4$, and $t=3$.

%\textbf{Food-101N:}
%We use ResNet-50 with ImageNet pre-trained weights, similar to Lee et al.~\shortcite{lee2017cleannet}. The models are evaluated on the test set from Food-101. All images are cropped at the center to a size of $224 \times 224 \times 3$. The network is trained with SDG with a momentum of $0.9$ and a decay of $0.001$ for a total of $5$ epochs. The learning rate starts at $0.01$ and is decreased to $0.005$, $0.002$, $0.001$ and $0.0002$ at epochs $2,3,4,5$, respectively. The batch size is set to $200$.\\

\vspace{.5\baselineskip}\noindent
\textbf{WebVision:}
We follow the setup in~\cite{li2020dividemix,chen2019understanding,jiang2020beyond} and use inception-resnet v2~\cite{szegedy2016inception}. The methods are evaluated on the provided validation set of the first 50 classes of the Google image subset. We resize the images to $320 \times 320$ and then random crop them to $299 \times 299$.
The network is trained with SDG with a momentum of $0.9$ and a weight decay of $0.0005$ for a total of $90$ epochs. 
The learning rate starts at $0.01$ and is decreased to $0.001$ at the 40th epoch and to $0.0001$ at the 80th epoch. We set $M=10$, $K=10$, and $t=2$.

\vspace{.5\baselineskip}\noindent
Furthermore, for the CIFAR-10 and CIFAR-100 datasets, we apply data augmentation techniques to the images in width and height shifts and random horizontal flips.\\

%\section{Additional Baselines}
%\textcolor{red}{The results for all the baselines can be added to a large table here?}

\end{document}